\documentclass{article}

 \usepackage[preprint]{neurips_2026}

\usepackage[utf8]{inputenc} % allow utf-8 input
\usepackage[T1]{fontenc}    % use 8-bit T1 fonts
\usepackage{hyperref}       % hyperlinks
\usepackage{url}            % simple URL typesetting
\usepackage{booktabs}       % professional-quality tables
\usepackage{amsfonts}       % blackboard math symbols
\usepackage{nicefrac}       % compact symbols for 1/2, etc.
\usepackage{microtype}      % microtypography
\usepackage{xcolor}         % colors
\usepackage{booktabs}
\usepackage{amsfonts}
\usepackage{amsmath}
\usepackage{amssymb}
\usepackage{nicefrac}
\usepackage{graphicx}
\usepackage{subcaption}
\usepackage{algorithm}
\usepackage{algorithmic}
\usepackage{multirow}
\usepackage{makecell}

\newcommand{\methodname}{MAPL}

\title{Learned Subspace Compression for Communication-Efficient Pipeline Parallelism}

\author{\parbox{\textwidth}{\centering
\vspace{0.75cm}
    Paul Janson$^{1, 2}$ \hspace{-10pt}
    \qquad  Edouard Oyallon$^{3}$
    \qquad Eugene Belilovsky$^{1,2}$ \vspace{5pt}\\
    \textnormal{{$^1$Concordia University ~~ $^2$Mila Quebec AI Institute ~~$^3$ CNRS, Sorbonne University}}
}}
\begin{document}

\maketitle

\begin{abstract}
Pipeline parallelism enables training of large language models that exceed single-device memory, yet inter-stage activation communication becomes the dominant bottleneck when trained on low-bandwidth networks. Recent work in this area has proposed using fixed orthogonal projections to compress activations. However, this still results in a significant performance degradation and requires a number of non-standard adaptations to constrain the optimization. A natural alternative is to learn a low rank projection for each pipeline stage, however maintaining the necessary orthogonality of these projectors during training remains a challenge. We present Manifold Aware Projection Learning (\textbf{MAPL}), a method that treats inter-stage compression as a learnable orthogonal projection under explicit Stiefel manifold (orthogonal matrices) constraints.  Rather than prescribing a fixed global subspace, MAPL lets each pipeline stage discover and continuously adapt its own task-optimal compression subspace via manifold-constrained steepest descent. To recover token-specific signals at stage boundaries, we introduce per-stage factorized anchor embeddings that allow for full-rank activation reconstruction with negligible communication overhead. 
We further show that we can incorporate residual vector quantization after projection with a streaming codebook synchronization protocol that amortizes dictionary communication.
Across LLaMA~\citep{llama1} models from 150M to 1B parameters we show that MAPL can be easily applied to the existing pipeline and can achieve high compression with neglibile performance degradation with a drastically improved tradeoffs in performance vs. compression compared to Subspace Networks.
% \footnote{Code is available at \href{https://anonymous.4open.science/r/learned-pipeline-2616/}{https://anonymous.4open.science/r/learned-pipeline-2616/}}

\end{abstract}

\section{Introduction}

% What is the problem 
Training large-scale foundation models across geographically distributed, heterogeneous hardware introduces communication challenges that centralized distributed training systems~\citep{rajbhandari2020zero} were not designed to handle~\citep{diloco,opendiloco,protocol_models,sparseloco,lidin2026covenant,nabli2025acco}. Centralized systems assume tightly coupled accelerator clusters with high-bandwidth interconnects and large memory for each accelerator, but real-world low-bandwith deployments often rely on lower resource accelerators as their backbone and must operate over commodity wide-area networks with limited bandwidth and memory. Since, modern models have grown well beyond the memory capacity of these individual accelerators, model parallelism is a practical necessity.  Pipeline parallelism~\citep{huang2019gpipe} responds by partitioning parameters across devices, allowing GPUs to easily host a few layers each. However, this introduces a new bottleneck that every micro-batch triggers activation exchanges across stages in both the forward and backward passes inducing a significant communication cost.
% When bandwidth is scarce, this communication cost dominates training, motivating \emph{low-bandwidth pipeline-parallel training} in which intermediate activations are aggressively compressed before transmission~\citep{protocol_models}.
\begin{figure}[!ht]
    \centering
    \includegraphics[width=0.7\linewidth]{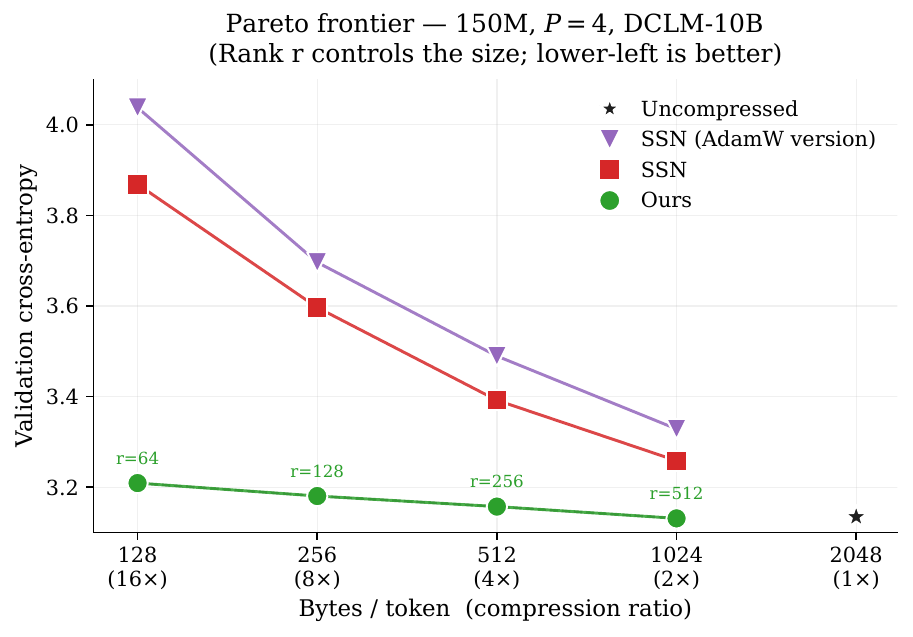}
    \caption{\textbf{Pareto frontier for compressed pipeline-parallel training.}
        Validation cross-entropy versus communication cost (bytes per token, with
        the corresponding compression ratio relative to the 2048-byte uncompressed
        baseline) for a 150M-parameter model trained with $P = 4$ pipeline stages on
        DCLM-10B  using the Muon optimizer~\citep{jordan2024muon} unless otherwise
stated. Lower-left is better. We compare our learned projection
        (``Ours,'' \textcolor{green}{green}) against uncompressed training (\textbf{$\star$}) and two SSN~\citep{protocol_models}
        baselines, SSN (AdamW~\citep{adam} version) and SSN. For all methods, the projection rank
        $r$ sets the activation size and hence the compression level
        ($r = 64, 128, 256, 512 \rightarrow 128\text{--}1024$ bytes/token). Our
        approach traces the Pareto frontier across all regimes, staying within
        ${\approx}0.08$ cross-entropy of uncompressed training even at $16\times$
        compression, whereas both SSN baselines degrade sharply as compression
        increases.}
    \label{fig:paretto-main}
\end{figure}

% Why is it important and interesting 
This communication cost motivates \emph{low-bandwidth pipeline-parallel training}, in which intermediate activations are aggressively compressed before transmission~\citep{protocol_models}. Beyond accessibility, compressing inter-stage activations is empirically well-motivated. There is growing evidence that foundation models converge to intrinsically low-rank solutions~\citep{janson2026stabilizingnativelowrankllm,protocol_models,yang2023spectral,galanti2025sgd}, suggesting that the full-rank activation tensors exchanged between pipeline stages carry significant redundancy. Exploiting this structure allows us to eliminate communication overhead and transmit only the information that actually drives learning. Activation compression, however, is fundamentally harder than compression in data-parallel training~\citep{sparseloco,powersgd}, where compressed gradients are computed independently across identical model replicas, compression errors can average out across workers, and distortions remain largely decoupled from the forward computation. In pipeline parallelism, stages hold complementary, non-overlapping subsets of the model, so the compressed activation produced by one stage becomes the direct input to the next; any transmission distortion therefore propagates through subsequent forward layers and backward gradients, accumulating across stages and potentially corrupting the learning signal even when the artifact is small.

% What is wrong with previous method SSN ? How's mine is different ?  

Recent work on Subspace Networks (SSNs)~\citep{protocol_models} addresses this by constraining each activation of width $d$ to a fixed, shared low-rank subspace of rank $r$ defined by an orthogonal matrix $U_r \in \mathbb{R}^{d \times r}$. Specifically, they transmit activations $h \in \mathbb{R}^d$ as low-dimensional coordinates $z = U_r^\top h$ and reconstruct them locally as $\hat{h} = U_r z$. Notably \cite{protocol_models} does not strictly constrain the activations to lie in the subspace thus the fixed projections cannot be removed at test time without performance degradation (See \S \ref{apdx:uk_ablation}). In this sense SSNs represent an architectural modification. 
While SSNs confirm that architectural modification is a viable design space for communication-efficient training, the approach requires intrusive constrained optimization: model weights are restricted to a common low-rank subspace, a modified AdamW~\citep{adam,adamw} optimizer is needed to maintain a weight in the subspace, and a static embedding offset is applied at each stage (See \S\ref{apx:subspace_background}). Consequently, we observe substantial accuracy degradation under a token-matched setting relative to uncompressed baselines.

% binding all layers to a \emph{common} subspace $U_k$ enforces a uniform representations that conflicts with the well-established observation that different layers encode qualitatively distinct features~\citep{zeiler2014visualizing}. 
% We see substantial degradation in accuracy for this method under token-matched setting relative to uncompressed baselines.  

% What are the key components of my approach and results?
Rather than prescribing a global basis for all layers, we let each pipeline stage \emph{learn} its own communication subspace jointly with the task objective. Intermediate activations exhibit substantial low-rank structure even under modern optimizers~\citep{jordan2024muon,kimimuon}, suggesting that communication-efficient representations can naturally emerge end-to-end. However
naively learning these projectors degrades the performance because standard gradient updates carry projectors off the Stiefel manifold (the set of all matrices with orthonormal columns), destroying the orthonormality on which isometric compression depends. Once a projector escapes the manifold, the model begins encoding features outside the intended subspace, leading to severe performance degradation. Crucially, we observe that these unconstrained models underperform even fixed orthogonal baselines, confirming that naive learning without manifold awareness is strictly worse than no learning at all (see \S\ref{apx:manifold_escape}). We identify this manifold escape as the principal failure mode and address it using manifold-constrained steepest-descent updates~\citep{yang2026manifold}, which keep projectors strictly on the Stiefel manifold throughout training. Furthermore, removing the global subspace eliminates the need for SSN-style embedding decompositions. Instead, we introduce \emph{factorized anchor embeddings}~\citep{Lan2020ALBERT:}, a low-rank factorization $E = E_p^{\mathrm{small}}P_p$ with fully trainable matrix $ E_p^{\mathrm{small}} \in \mathbb{R}^{V \times r}$ and a frozen matrix $P_p \in \mathbb{R}^{r \times d}$, where $V$ is the vocabulary size. This approach keeps the parameter count low while permitting the effective embedding to recover its full rank at each stage. Finally, we further reduce bandwidth via vector quantization (VQ) on the low-rank manifold, employing a lightweight dictionary-stream protocol that amortizes codebook synchronization across many activation exchanges.

To this end, we introduce \textbf{Manifold Aware Projection Learning (MAPL)}, which treats inter-stage communication as a learnable geometric projection rather than a fixed architectural constraint. We make the following contributions. We show that enforcing a shared global subspace across pipeline stages degrades learning in token-matched settings and instead propose allowing each stage to learn its own compression subspace on the Stiefel manifold, which we find is better suited to low-bandwidth activation transmission (\S~\ref{sec:method_construction}). Second, we introduce a low-rank embedding factorization that replaces the fixed token embedding decomposition used in SSN-style approaches with stage-specific learnable embeddings. Combined with vector quantization and a streaming codebook synchronization protocol, this design further reduces inter-stage bandwidth while requiring only integer token IDs to be transmitted between stages (\S~\ref{sec:method_vq}). Third, we evaluate MAPL on models from 150M to 1B parameters, where it recovers downstream performance to within 1\% of the uncompressed baseline in token-matched settings and outperforms SSNs by 5\% across all evaluated scales (\S~\ref{results:main}). As Fig.~\ref{fig:paretto-main} shows, MAPL traces the Pareto frontier of cross-entropy versus compression, improving the achievable tradeoff over prior methods.

\section{Related works}

\textbf{Low Bandwidth Pre-training.}
The growing utility of large AI systems has motivated efforts to democratize large-scale model training across decentralized bodies of participants connected over bandwidth-constrained networks such as the open internet. Early works on decentralized optimization established theoretical foundations: \citep{lian2017can} showed that decentralized SGD can match centralized convergence rates, while subsequent analyses unified gossip-based optimization with compressed communication and local updates~\citep{koloskova2019decentralized,koloskova2020unified}. To make distributed training communication-efficient, a rich line of work explored gradient compression through sparsification~\citep{wang2017efficient,wangni2018gradient,lin2018deep}, scalar and sign-based quantization~\citep{alistarh2017qsgd,bernstein2018signsgd,tang20211,dettmers2022bit}, low-rank approximation~\citep{vogels2019powersgd}, and error-feedback mechanisms that correct biased compressors~\citep{wu2018error,karimireddy2019error}. Building on these foundations, volunteer-style collaborative training was demonstrated by DeDLOC~\citep{diskin2021distributed} and CocktailSGD~\citep{wang2023cocktailsgd} for training over slow networks. Recent work has focused on data-parallel pre-training over the internet: DiLoCo-style methods~\citep{diloco,sparseloco,opendiloco,douillard2025streaming} adapt federated-averaging~\citep{local_sgd} to drastically reduce synchronization frequency. DeMo~\citep{demo} exploits fast orthonormal transforms, top-k sparsification, and error feedback to enable multi-datacenter training. ACCO~\citep{acco} overlaps the synchronization of delayed gradients with the computation of new gradients to increase GPU utilization. SparseLoCo~\citep{sparseloco} unifies error feedback, sparsification, and quantization to achieve internet-scale decentralized training~\citep{lidin2026covenant}. However, all of these methods assume that each accelerator can hold a complete model replica. Pipeline-parallel training exhibits fundamentally different dynamics from DDP: SWARM~\citep{ryabinin2023swarm} identified the square-cube law, showing that computation scales cubically while communication scales quadratically, and prior work explored decentralized model parallelism via mixture-of-experts routing~\citep{ryabinin2020towards} and reversible architectures that decouple forward and backward passes across stages~\citep{rivaud2025petra}. Several works expanded on compressing activations~\citep{chen2021actnn,bian2024does} and the change in activations~\citep{wang2022fine}. Subspace Networks~\citep{protocol_models} exploited the empirical rank collapse of transformer residual streams to project boundary activations onto a fixed orthonormal basis. While effective, this imposes a single representational space across all transformer layers. In our work, we challenge this assumption to learn specialized, per-stage compression.

\textbf{Efficient language models.}
A parallel line of work has investigated low-rank geometric structure for memory-efficient gradient compression and model compression, motivated by the empirical observation that deep networks converge to low-rank solutions~\citep{lecun1989optimal,yu2017compressing,martin2021implicit,yang2023spectral,frankle2018the,galanti2025sgd}. GaLore~\citep{zhao2024galore} style methods~\citep{chen2026fira,lialin2024relora} projected stochastic gradients onto low-rank subspaces to reduce optimizer state memory, enabling full-parameter pre-training under tight memory budgets. Several works~\citep{janson2026stabilizingnativelowrankllm,mo2025parameter,wei2024_building} further explored training low-rank factorized weights to reduce memory and communication requirements. LTE~\citep{huh2024training} used parallel LoRA-style adapters to pre-train models from scratch with reduced memory and bandwidth across parallel accelerators. Pufferfish~\citep{wang2021pufferfish} modified the architecture to obtain trained factorized layers that reduce communication costs, but focused on data parallelism. Our work similarly modifies the transformer architecture, but with the complementary goal of enabling efficient inter-stage activation communication for pipeline parallelism.

% =====================================================================
\section{Manifold Aware Projection Learning}
\label{sec:method}
% =====================================================================

\begin{figure}[!t]
    \centering
    \includegraphics[width=0.99\linewidth]{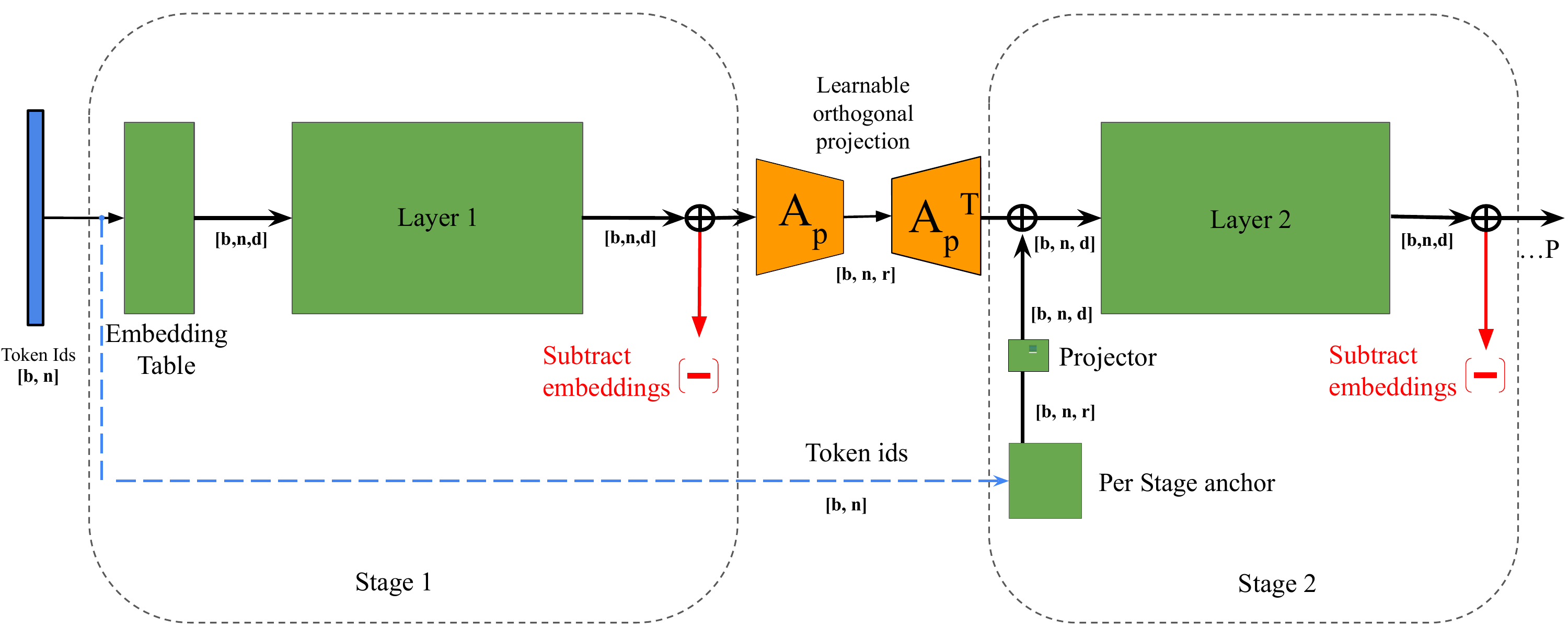}
    \caption{\textbf{Overview of MAPL compression at a pipeline stage boundary, 
repeated across all $P-1$ inter-stage boundaries.} At each boundary, the 
token-dependent offset is subtracted from the boundary activation 
$X^{b_p} \in \mathbb{R}^{b \times n \times d}$ before transmission (\textcolor{red}{red} arrow): 
the first stage subtracts the original token embeddings, while subsequent stages 
subtract their per-stage factorized anchor embeddings 
$E_p^{\mathrm{small}}[t_{ids}]\, P_p$, where $P_p$ is 
a frozen orthogonal projector. The residual is then projected to a low-dimensional 
representation $Z^{b_p} \in \mathbb{R}^{b \times n \times r}$ via a learnable 
orthogonal projector $A_p \in \mathrm{St}(d, r)$, reducing inter-stage communication 
volume by a factor of $d/r$. At the receiving stage, $A^\top_{p}$ reconstructs the 
full-dimensional activation, to which the destination-stage factorized anchor is 
added to restore the offset. Integer token IDs are transmitted alongside the 
compressed activation on the same channel at negligible cost (\textcolor{blue}{blue} dashed line), enabling each stage to look up its local anchor without additional 
bandwidth overhead.}
    \label{fig:method_overview}
\end{figure}

Figure~\ref{fig:method_overview} summarizes MAPL. At each of the inter-stage boundaries, We subtract the token embeddings and project the boundary activations to a low-dimensional subspace via a learnable orthogonal projector, transmitted, and reconstructed at the receiving stage. We derive this design bottom-up — beginning from an empirical observation about the structure of boundary activations (\S\ref{sec:method_obs}), which motivates the construction (\S\ref{sec:method_construction}) and its optimization procedure (\S\ref{sec:method_spel}). We validate each component ablatively in \S\ref{sec:method_validation} and sketch a composable vector-quantization extension in \S\ref{sec:method_vq}.

\subsection{Boundary activations are intrinsically low-rank}
\label{sec:method_obs}
\begin{figure}[t]
    \centering
    \includegraphics[width=0.99\textwidth]{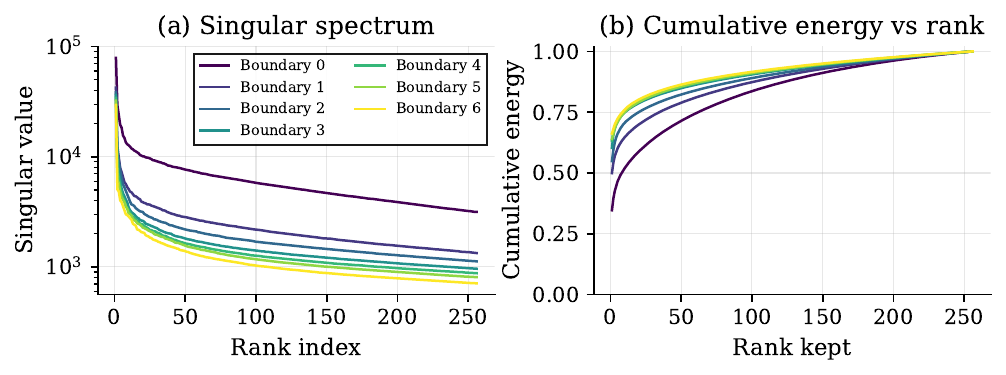}
    \caption{\textbf{Boundary activations exhibit intrinsic low-rank structure across all pipeline stages, with rank-250 truncation retaining ${\geq}99\%$ of activation energy.} (a) Singular value spectra of the centered boundary activations \(X^{b_p} - E[t_{\text{ids}}]\)
(reshaped to \((B \cdot T) \times d\)) for all \(P-1 = 7\) inter-stage boundaries of a 150M LLaMA model (\(d=1024\), \(P=8\)) trained with Muon~\citep{jordan2024muon} on DCLM~\citep{dclm}. 
The \(x\)-axis indexes singular values in descending order; the \(y\)-axis is on a log scale. 
Each colored line corresponds to a distinct pipeline boundary. 
(b) Cumulative energy \(\frac{\sum_{i=1}^{r}\sigma_i^2}{\sum_{i=1}^{d}\sigma_i^2}\) as a function of retained rank \(r\). 
Color coding is shared across both panels.}
    \label{fig:c1}
\end{figure}
% ---------------------------------------------------------------------

\textbf{Setup} Pipeline parallelism partitions the $L$ layers of a decoder-only Transformer~\citep{touvron2023llama} into $P$ contiguous stages $\mathcal{S}_1, \dots, \mathcal{S}_P$, each residing on a dedicated device. Let $E$ denote the embedding table and $t_{\mathrm{ids}}$ the input token indices. At each inter-stage boundary $p \in \{1, \dots, P{-}1\}$, the forward pass communicates the boundary activation $X^{b_p} \in \mathbb{R}^{B \times T \times d}$, while the backward pass transmits its gradient; $B$, $T$, and $d$ denote batch size, sequence length, and hidden dimension, respectively.

\textbf{Motivating example} We profile a 150M-parameter LLaMA model~\citep{llama1} trained for 3{,}000 steps on DCLM~\citep{dclm} with the Muon optimizer~\citep{jordan2024muon} at $P{=}8$ pipeline stages. To isolate representational structure of the residual stream, we subtract the token embeddings and analyze the residual $X_{res} = X^{b_p} - E[t_{\mathrm{ids}}]$, reshaped to $(B{\cdot}T) \times d$, via singular value decomposition (SVD). We quantify spectral concentration through the cumulative energy ratio:
\[
\mathcal{E}(r) = \frac{\sum_{i=1}^{r} \sigma_i^2}{\sum_{i=1}^{d} \sigma_i^2}.
\]
 As shown in Figure~\ref{fig:c1}, boundary activations exhibit pronounced low-rank structure across all pipeline stages: a rank of $r \approx 250$ suffices to retain ${\geq}99\%$ of the total activation energy, despite the ambient dimension $d{=}1024$. The inter-stage signal therefore concentrates near a submanifold of effective dimension roughly one quarter of the full representation space.

This finding extends~\citep{protocol_models}, which observed rank collapse in projection matrices and responded by explicitly constraining weight rows to a shared low-rank subspace. Crucially, our setting requires \emph{no} such contraint on weights: the low-rank structure emerges organically during training, without regularization or weight constraints. We therefore define a rank-$r$ projector ($r < d$) as \emph{information-preserving} for this signal, and argue that a learned projection can \emph{discover}---rather than impose---the latent subspace the activations already inhabit, enabling principled compression of inter-stage communication.

% ---------------------------------------------------------------------
\subsection{Compression via per-stage learned orthogonal projectors}
\label{sec:method_construction}
% ---------------------------------------------------------------------

We introduce, at each pipeline-stage boundary $p$, a learnable orthogonal projector $A_p$ that lies on the Stiefel manifold
$$
\mathrm{St}(d, r) = \{A \in \mathbb{R}^{d \times r} : A^\top A = I_r\}.
$$
We equip each boundary with a per-stage anchor $E_p^{\mathrm{small}}[t_{\mathit{ids}}]$ and a companion projector $P_p$ that absorbs the high-rank, token-driven offset of the residual stream. Forward compression and reconstruction then proceed as
\begin{equation}
    Z^{b_p} = \bigl(X^{b_p} - E_p^{\mathrm{small}}[t_{\mathit{ids}}] P_p \bigr)\, A_p
        \;\in\; \mathbb{R}^{B \times T \times r},
        \label{eq:project}
\end{equation}
\begin{equation}
    \hat{X}^{b_p} = Z^{b_p}\, A_p^\top + E_{p+1}^{\mathrm{small}}[t_{\mathit{ids}}] P_{p+1}
        \;\in\; \mathbb{R}^{B \times T \times d}.
        \label{eq:reconstruct}
\end{equation}
By enforcing the Stiefel constraint, we guarantee that $A_p^\top$ is the exact inverse of $A_p$. Projection and reconstruction are therefore isometric on the column space of $A_p$ and the method achieves an exact compression ratio of $r/d$. In the backward pass we symmetrically route the gradient $\partial \mathcal{L}/\partial X^{b_p}$ through the same orthogonal projector $A_p$.

The rank-$r$ projector $A_p$ captures the dominant low-rank component of the boundary activation $X^{b_p}$ while deliberately leaving the token-frequency offset unaddressed; this offset is inherently high-rank and would otherwise exhaust projector capacity. In SSN~\citep{protocol_models}, the embedding table is decoupled into static high rank and learnable low rank components. Each stage added a static high rank offset as the learnable token embeddings forced to be in the same subspace as the weights~\citep{protocol_models}. We instead add a learnable offset. To reduce the parameter pressure at each stage, we factorize the offset as
\begin{equation}
    E_p^{\mathrm{small}}[\mathit{ids}]\, P_p,
    \qquad
    E_p^{\mathrm{small}} \in \mathbb{R}^{V \times r},
    \qquad
    P_p \in \mathbb{R}^{r \times d},
    \label{eq:anchor}
\end{equation}
where $E_p^{\mathrm{small}}$ is a trainable embedding table and $P_p$ is a fixed random orthonormal matrix. Only the integer token IDs cross the communication channel for this reconstruction. Consequently, the anchor adds negligible bandwidth overhead. 
% We detail the choice of the anchor dimension $k$, the associated parameter overhead at scale, and ID-packing implementation in Appendix~\ref{app:impl}.

% ---------------------------------------------------------------------
\subsection{Optimization via SPEL}
\label{sec:method_spel}
% ---------------------------------------------------------------------

We update each $A_p$ jointly with the model weights via SPEL (Spectral
Steepest Descent on the Stiefel Manifold)~\citep{yang2026manifold},
using the task loss as the only signal. Given the Euclidean gradient
$g_t = \partial \mathcal{L} / \partial A_p$ at step $t$ and learning rate $\alpha$, update consists of:
\begin{align}
    g_t^R &= g_t - A_p\, \mathrm{sym}(A_p^\top g_t),
        && \text{(tangent projection)} \label{eq:tangent} \\
    m_t   &= \beta\, m_{t-1} + (1 - \beta)\, g_t^R,
        && \text{(heavy-ball momentum)} \label{eq:momentum} \\
    d_t   &= \mathrm{PolarExpress}(m_t),
        && \text{(spectral LMO direction)} \label{eq:lmo} \\
    A_p   &\gets \mathrm{PolarExpress}\bigl(A_p - \alpha\, d_t\bigr),
        && \text{(retraction onto $\mathrm{St}(d,r)$)} \label{eq:retract}
\end{align}
where $\mathrm{sym}(M) = \tfrac{1}{2}(M + M^\top)$. Polar express~\citep{amsel2025polar} calculates the spectral norm LMO direction which is a similar parameter update as Muon~\citep{jordan2024muon} and retracts the matrix back to the stiefel manifold. SPEL inherits the
$\mathcal{O}(1/\sqrt{T})$ convergence rate of first-order methods on
smooth manifolds~\citep{yang2026manifold}. We send the updated $A_p$ to the next stage after every optimizer step. This is a minimal communication cost when compared with activationa and gradient communication. 

Unlike SSN's Grassmann update that refresh every ${\sim}500$
steps~\citep{protocol_models}, SPEL updates $A_p$ at every step, letting
the subspace track the evolving activation geometry. In practice we choose a learning rate an order smaller than parameter updates ($\times0.1$). Algorithm~\ref{alg:MAPL} summarizes the full
per-boundary update.

\begin{algorithm}[t]
\caption{MAPL: per-boundary compression and projector update.}
\label{alg:MAPL}
\begin{algorithmic}[1]
\REQUIRE Boundary activation $X^{b_p}$, projector $A_p \in \mathrm{St}(d,r)$,
        momentum buffer $m_p$, anchor $E_p^{\mathrm{small}}$, companion projector $P_p$,
        token IDs $t_{ids}$, projector learning rate $\alpha$, momentum $\beta$.
\vspace{0.3em}
\STATE \textbf{Project (sender):}\quad
        $Z^{b_p} \gets \bigl(X^{b_p} - E_p^{\mathrm{small}}[t_{ids}]\, P_p \bigr)\, A_p$
\STATE \textbf{Transmit} $Z^{b_p}$ and $t_{ids}$
\STATE \textbf{Reconstruct (receiver):}\quad
        $\hat{X}^{b_p} \gets Z^{b_p}\, A_p^\top + E_{p+1}^{\mathrm{small}}[t_{ids}]\, P_{p+1}$
\vspace{0.3em}
\STATE \textbf{Projector update (SPEL):}
\STATE \quad $g \gets \partial\mathcal{L}/\partial A_p$
\STATE \quad $g_R \gets g - A_p\,\mathrm{sym}(A_p^\top g)$
        \hfill\COMMENT{tangent projection}
\STATE \quad $m_p \gets \beta\, m_p + (1-\beta)\, g_R$
        \hfill\COMMENT{momentum}
\STATE \quad $d \gets \mathrm{PolarExpress}(m_p)$
        \hfill\COMMENT{LMO direction}
\STATE \quad $A_p \gets \mathrm{PolarExpress}(A_p - \alpha\, d)$
        \hfill\COMMENT{retraction}
\end{algorithmic}
\end{algorithm}

\begin{figure}
    \centering
    \includegraphics[width=0.99\linewidth]{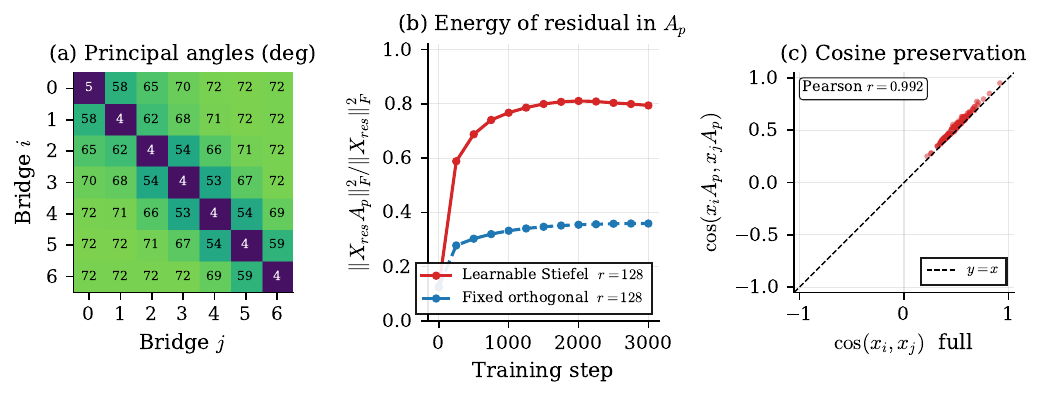}
    \caption{\textbf{Empirical validation of learned orthogonal projectors for activation compression in pipeline-parallel training.} (a) Pairwise mean principal angles (degrees) between learned Stiefel manifold projectors across pipeline stages $i$ and $j$, computed via $\arccos(\sigma_k(A_i^\top A_j))$. Large off-diagonal angles (up to $72^\circ$) confirm that projectors across non-adjacent stages converge to geometrically distinct subspaces, while near-zero diagonal values verify self-consistency. (b) Frobenius-norm energy of residual activations preserved under projection, $\|X_{\text{res}} A_p\|_F^2 / \|X_{\text{res}}\|_F^2$, as a function of training step for learned Stiefel ($r=128$, red) versus fixed orthogonal ($r=128$, blue) projectors. Learned projectors retain approximately $2.2\times$ more residual energy ($\sim0.80$ vs. $\sim0.36$), demonstrating superior task-adaptive compression. (c) Cosine similarity preservation between token pairs: $\cos(x_i A_p,\, x_j A_p)$ versus $\cos(x_i, x_j)$ evaluated on the full-dimensional representations. A Pearson correlation of $r = 0.992$ confirms that the learned projection $A_p$ faithfully preserves the pairwise relational geometry of the activation space, closely tracking the identity line ($y = x$).}
    \label{fig:combined}
\end{figure}
% ---------------------------------------------------------------------
\subsection{Empirical Validation}
\label{sec:method_validation}

We validate three properties of our method
(\S\ref{sec:method_construction}) and show learning the projection is naturally more favorable compression than having a fixed projector. We show that per-boundary
projectors discover geometrically distinct subspaces for each different stage. The learned $A_p$
captures substantially more activation energy than a fixed orthonormal basis
of equal rank, and the learned projection faithfully preserves pairwise
token similarity after compression.\\
\textbf{Each stage learns a distinct subspace.}
Figure~\ref{fig:combined} (a) reports pairwise mean principal angles between the
seven learned projectors of all boundaries, computed as $\bar{\theta} = \frac{1}{r}\sum_{k=1}^{r}\arccos\bigl(\sigma_k(A_i^\top A_j)\bigr)$. Off-diagonal angles range from $53^\circ$ to $72^\circ$, with
adjacent stages exhibiting the greatest overlap and distant stages approaching
near-orthogonality. This change in subspace alignment is consistent with the
residual stream transitioning from lexical to task-specific representations
across depth~\citep{zeiler2014visualizing}.\\
% Tying projectors across stages---the natural
% baseline---sacrifices $0.04$ validation nats (Table~\ref{tab:ssn_compare}),
% confirming that each pipeline boundary warrants its own $A_p$.
\textbf{The learned projector captures substantially more energy.}
Figure~\ref{fig:combined} (b) compares the residual energy retained under
projection, $\|X_{\mathrm{res}} A_p\|_F^2 / \|X_{\mathrm{res}}\|_F^2$,
between a fixed random orthonormal basis and the jointly trained $A_p$ at
matching rank $r{=}128$. Note that $\|A\|_F^2 = \sum_{i} \sigma_i^2$ . The fixed basis plateaus near $36\%$ of captured energy while the learned
projector reaches ${\sim}80\%$ within $1{,}500$ steps and continues to
improve---a factor-of-two gain in effective rank utilization. Inspection of
weight stable rank confirms that the preceding attention out projection and MLP projection layers\\
adapt to this space defined by $(A_p)$ (Appendix~\ref{apx:rank_collapse}). We see that the compression actively induces the low rank structure. \\
% The compression does not
% merely exploit pre-existing low-rank structure; it actively induces it.
\textbf{The learned projection preserves pairwise token geometry.}
Figure~\ref{fig:combined} (c) plots the cosine similarity between projected
token pairs, $\cos(x_i A_p,\, x_j A_p)$, against their full-dimensional
counterpart $\cos(x_i, x_j)$. The Pearson correlation of $r = 0.992$ with the
identity line demonstrates that $A_p$ behaves as a near-isometry over the
empirical token distribution and the relative angular structure in the activation
space is preserved under compression. This property is not enforced
explicitly and it emerges from Stiefel optimization alone and provides
geometric evidence that the transmitted representation retains the relational
information downstream stages require for computation.

% ---------------------------------------------------------------------
\subsection{Composing with vector quantization}
\label{sec:method_vq}
% ---------------------------------------------------------------------
To minimize communication overhead during pipeline-parallel training, we employ Multi-Codebook Vector Quantization (MCVQ) for compressing both forward activations and backward gradients. 
The method operates on a low-rank projected representation, decomposed into $G$ groups, and applies $R$ rounds of residual quantization to achieve high compression ratios while preserving representational fidelity. \\
\textbf{Quantization Scheme} Formally, the projected representation is compressed via residual vector quantization using a per-stage codebook. 
$Z^{b_p} \in \mathbb{R}^{B \times T \times r}$
is compressed using a per-stage codebook $\mathcal{C}_p \in \mathbb{R}^{r \times K}$.
Each stage quantizes the residual error from the previous round, enabling progressive refinement of the compressed representation across $R$ rounds.\\
\textbf{Codebook Synchronization}
Sender and receiver nodes synchronize $\mathcal{C}_p$ through a streaming dictionary update protocol: the codebook is partitioned into random subsets, with a $\frac{1}{K}$ fraction of codes transmitted per micro-batch. 
This design is motivated by the empirical observation of \citep{zhang2024codebook} that VQ codebooks evolve slowly over the course of training, ensuring that the staleness introduced by incremental updates remains negligible with respect to overall bandwidth and convergence.

% ----------------------------------------------------------------------------
% \subsection{Muon Optimizer}
% \label{sec:bg_muon}
% % ----------------------------------------------------------------------------

% Muon~\citep{jordan2024muon} applies spectral-norm steepest descent to weight updates: each gradient is orthogonalized via Newton-Schulz iterations to produce a near-orthogonal update matrix. This is equivalent to solving the LMO over the spectral-norm ball, yielding updates whose singular values are approximately 1. Muon has demonstrated superior performance to AdamW for LLM pretraining for 2D weight matrices.

% Critically, Muon's update rule is \emph{not} compatible with SSN's subspace confinement strategy. SSN requires row-wise constant adaptive scaling to keep $W_{p2}$ rows in $\text{span}(U_k)$; Muon's spectral orthogonalization applies a global transformation that does not preserve subspace membership. Our approach sidesteps this incompatibility by decoupling the projector optimization (SPEL on $\St(d, r)$) from the model weight optimization (Muon or any other optimizer).

% Recent work~\citep{numuon} has shown that Muon-trained networks also exhibit rank collapse in their weight matrices, suggesting that the low-rank activation structure exploited by SSN is not specific to AdamW but is a more general property of Transformer optimization dynamics.

% ============================================================================

% =====================================================================
% =====================================================================

\section{Experiments}
\label{sec:experiments}
% =====================================================================
% ----------------------------------------------------------------------
\subsection{Experimental Setup}
\label{sec:setup}
% ----------------------------------------------------------------------
\textbf{Architecture and scales.} We evaluate our method on decoder-only transformers based on the LLaMA~\citep{llama1} architecture across three parameter scales: 150M, 500M, and 1B. All models employ a context length of 2048 tokens and are trained with the LLaMA tokenizer (vocabulary size 32,000) in bf16 precision. All models are pre-trained on the DCLM corpus~\citep{dclm}, with 5M tokens held out uniformly at random as a validation set. Following the Chinchilla compute-optimal scaling regime~\citep{hoffmann2022training}, each model is trained on a token budget of 20 tokens per parameter. Implementation details are provided in Appendix~\ref{apx:implementation_details}. All the methods are compared at equal token budgets.

\textbf{Optimization}
We adopt a hybrid optimizer configuration in which the 2D hidden weight matrices are updated using Muon~\citep{jordan2024muon}, while embeddings, biases, and the output projection are updated with AdamW~\citep{adamw}. The Muon learning rate is set to $\eta_\mu = 0.02$ for the 150M, 500M sizes and $\eta_\mu = 0.01$ for the 1B case, and the AdamW learning rate is coupled to it via a multiplier of $0.5$, yielding $\eta_{\text{adam}} = 0.5 \cdot \eta_\mu$. Training uses a global batch size of 512. We choose the multiplier of $0.1$ for learning rate we use to update the projector using \S~\ref{sec:method_spel}.  Our primary experiments are conducted under pipeline-parallel configurations with $P \in \{4, 8\}$ stages.

\textbf{Baselines.} We compare against methods while holding all shared hyperparameters fixed to ensure a fair comparison. \textbf{Uncompressed} transmits activations and gradients across pipeline boundaries at the full model width and therefore serves as an upper bound on attainable quality. \textbf{SSN}~\citep{protocol_models} projects inter-stage tensors onto a learned low-rank subspace(\S~\ref{apx:subspace_background}). \textbf{SSN (AdamW version)} use SSN with AdamW optimizer for all the parameters. \textbf{\methodname} applies our proposed compression scheme to inter-stage activations and gradients. Finally, \textbf{\methodname + VQ} augments our approach with vector quantization of the projected representation, effectively doubling the compression ratio.

\textbf{Evaluation}
We report cross-entropy validation loss on the held-out DCLM split as our primary metric, together with the relative degradation $\Delta\%$ with respect to the Uncompressed baseline at the same scale. We additionally evaluate zero-shot downstream accuracy on HellaSwag~\citep{zellers2019hellaswag}, PIQA~\citep{bisk2020piqa}, ARC-Easy, and ARC-Challenge~\citep{clark2018think} using \texttt{lm-evaluation-harness}~\citep{eval-harness}. We also report average accuracy across all the tasks.

\begin{table}[t]
\centering
\footnotesize
\begin{tabular}{@{}llrrrrrr@{}}
\toprule
\textbf{Scale} & \textbf{Method} & \textbf{Bytes/token} & \textbf{Comp.} & \textbf{P=4 Loss} & \textbf{P=4 $\Delta$ \%} & \textbf{P=8 Loss} & \textbf{P=8 $\Delta$ \%} \\ \midrule
150M           & \textit{Uncompressed}                    & 2048 & ---  & 3.13  &          ---  & 3.13  &           ---  \\
150M           & SSN~\citep{protocol_models} (AdamW~\citep{adam} version)& 512  & 4$\times$    & 3.49  & 11.49\%  & 3.52  & 12.46\%   \\
150M           & SSN~\citep{protocol_models}                              & 512  & 4$\times$    & 3.39  & 8.37\%   & 3.40  & 8.63\%    \\
150M           & \methodname                     & 512  & 4$\times$    & 3.156 & \textbf{0.84\%} & 3.165 & \textbf{1.11\%} \\
150M           & \methodname + VQ                & 256  & 8$\times$    & 3.165 & 1.11\%   & 3.170 & 1.28\%    \\ \midrule
500M           & \textit{Uncompressed}                    & 3072 & ---  & 2.84  &          ---  & 2.84  &           ---  \\
500M           & SSN~\citep{protocol_models}  (AdamW~\citep{adam} version)             & 512  & 6$\times$    & 3.25  & 14.34\%  & 3.29  & 15.90\%   \\
500M           & SSN~\citep{protocol_models}                              & 512  & 6$\times$    & 3.09  & 8.92\%   & 3.12  & 9.90\%    \\
500M           & \methodname                     & 512  & 6$\times$    & 2.79  & \textbf{-1.90\%} & 2.84 & \textbf{0.00\%} \\
500M           & \methodname + VQ                & 256  & 12$\times$   &       2.92&          2.75\%& 2.88  & 1.49\%    \\ \midrule
1B             & \textit{Uncompressed}                    & 4096 & ---  & 2.68  &          & 2.68  &           \\
1B             & SSN~\citep{protocol_models}  (AdamW~\citep{adam} version)             & 512  & 8$\times$ &       3.38&          26.34\%& 3.39  & 26.42\%   \\
1B             & SSN~\citep{protocol_models}                              & 512  & 8$\times$ &       3.05&          13.93\%& 3.08  & 15.05\%   \\
1B             & \methodname                     & 512  & 8$\times$    &       2.72&          \textbf{1.38\%}& 2.73  & \textbf{2.02\%} \\
1B             & \methodname + VQ     & 256   & 16$\times$  & 2.76  &   3.01\%    &    2.74      &    2.30\%             \\ \bottomrule
\end{tabular}
\vspace{1em}
\caption{\textbf{MAPL closes the performance gap to uncompressed training across all model scales while delivering 4--16$\times$ communication compression.} Cross entropy validation loss and compression results for 150M, 500M, and 1B parameter models under pipeline-parallel settings with P=4 and P=8 show that both MAPL  and its vector-quantized variant consistently match or approach the uncompressed baseline. In contrast, they substantially outperform the SSN baselines in accuracy. Best uncompressed values are marked with \textbf{bold} text}
\label{tab:my-table}
\end{table}

\subsection{Main Results}
\label{results:main}

\textbf{\methodname{} closes the gap to the uncompressed baseline across all scales,} Table~\ref{tab:my-table} shows that, at $4\times$ compression on the 150M model, \methodname{} achieves validation losses of $3.156$ ($P{=}4$) and $3.165$ ($P{=}8$). This is just $0.84\%$ and $1.11\%$ 
above the uncompressed reference of $3.13$. At this scale, we can see that SSN degrades much higher in performance in both pipeline cases of pipeline split.
Scaling to 500M parameters under $6\times$ compression, \methodname{} tracks the baseline within $1.90\%$ 
($P{=}4$) and $0.08\%$ ($P{=}8$); at 1B under $8\times$ compression, it holds losses of $2.72$ and $2.73$, degrading by only $1.38\%$ and $2.02\%$ 
respectively. At all scales the SSN performs much worse and looses a maximum of nearly 14\% performance compared to uncompressed baseline. 

We attribute the stability of \methodname{} across model scales to the per-stage Stiefel-constrained projectors, which faithfully capture the intrinsic low-rank geometry of the activations identified in \S\ref{sec:method_obs}. 
Composing \methodname{} with residual vector quantization (\S\ref{sec:method_vq}) doubles the compression ratio while incurring only marginal additional loss. 
At 1B parameters, \methodname{}+VQ achieves $16\times$ inter-stage compression at $3.01\%$ ($P{=}4$) and $2.30\%$ ($P{=}8$) degradation. 
We trace this behavior to the empirical near-isometry of the learned projector (Figure~\ref{fig:combined}c).By preserving pairwise similarity on the low-rank manifold, the codebook in $\mathbb{R}^r$ inherits a well-conditioned distribution 
that aligns itself naturally for vector quantization.

\textbf{\methodname{} preserves downstream performance on various tasks.} 
Validation-loss improvements translate directly to zero-shot performance as shown in Table~\ref{tab:downstream-accuracy}. 
Across the tested downstream tasks, \methodname{} closely tracks the uncompressed baseline at every scale. 
At 500M parameters, the average accuracy gap is $0.2$ points at $P{=}4$ ($41.8$ vs.\ $42.0$) and $0.4$ points at $P{=}8$ ($41.6$ vs.\ $42.0$); 
at 1B, the corresponding gaps are $0.8$ and $1.5$ average points. 
SSN baselines, by contrast, suffer accuracy drops as large as $8.8$ points at 1B. This shows that the learning capacity is restricted due to the global-subspace weight constraint and shows degraded performance at token matched comparisons.
The \methodname{}+VQ variant incurs a more pronounced performance loss in downstream tasks.

% ----------------------------------------------------------------------

% Please add the following required packages to your document preamble:
% \usepackage{booktabs}
% \usepackage{multirow} % (optional, not used here but good for future tweaks)

\begin{table}[htbp]
\centering
\tiny
\begin{tabular}{@{}llccccc|ccccc@{}}
\toprule
\multirow{2}{*}{\textbf{Size}} & \multirow{2}{*}{\textbf{Config}}
  & \multicolumn{5}{c|}{\textbf{$P{=}4$}}
  & \multicolumn{5}{c}{\textbf{$P{=}8$}} \\
\cmidrule(lr){3-7}\cmidrule(l){8-12}
 & & HellaSwag & PIQA & ARC-E & ARC-C & Avg
   & HellaSwag & PIQA & ARC-E & ARC-C & Avg \\
\midrule
\multirow{5}{*}{150M}
 & \textit{Uncompressed}                                              & 28.1 & 60.1 & 38.2 & 22.8 & 37.3 & 28.1 & 60.1 & 38.2 & 22.8 & 37.3 \\
 & SSN~\citep{protocol_models} (AdamW~\citep{adam} version) & 26.5 & 56.1 & 32.5 & 20.6 & 33.9 & 26.8 & 56.0 & 32.1 & 20.6 & 33.9 \\
 & SSN~\citep{protocol_models}                              & 27.6 & 56.7 & 34.4 & 20.9 & 34.9 & 27.0 & 56.3 & 33.8 & 21.7 & 34.7 \\
 & \methodname                                              & 28.0 & 59.4 & 37.0 & 22.6 & \textbf{36.7} & 27.8 & 59.3 & 37.8 & 21.5 & \textbf{36.6} \\
 & \methodname{} + VQ                                       & 28.4 & 52.8 & 31.7 & 22.7 & 33.9 & 26.4 & 50.5 & 29.1 & 25.9 & 33.0 \\
\midrule
\multirow{5}{*}{500M}
 & \textit{Uncompressed }                                             & 35.7 & 64.4 & 43.2 & 24.8 & 42.0 & 35.7 & 64.4 & 43.2 & 24.8 & 42.0 \\
 & SSN~\citep{protocol_models} (AdamW~\citep{adam} version) & 27.0 & 59.5 & 35.6 & 21.2 & 35.8 & 27.4 & 57.9 & 34.6 & 22.1 & 35.5 \\
 & SSN~\citep{protocol_models}                              & 29.1 & 59.8 & 38.5 & 22.3 & 37.4 & 28.5 & 59.1 & 37.2 & 22.2 & 36.7 \\
 & \methodname                                              & 35.1 & 64.0 & 43.6 & 24.3 & \textbf{41.8} & 34.5 & 64.7 & 42.4 & 24.7 & \textbf{41.6} \\
 & \methodname{} + VQ                                       & 25.9 & 50.8 & 26.6 & 26.7 & 32.5 & 26.9 & 51.2 & 27.2 & 24.3 & 32.4 \\
\midrule
\multirow{5}{*}{1B}
 & \textit{Uncompressed }                                            & 38.8 & 66.3 & 46.8 & 24.5 & 44.1 & 38.8 & 66.3 & 46.8 & 24.5 & 44.1  \\
 & SSN~\citep{protocol_models} (AdamW~\citep{adam} version)& 27.5 & 57.9 & 35.1 & 20.8 & 35.3 & 27.4 & 57.6 & 35.6 & 20.7 & 35.3 \\
 & SSN~\citep{protocol_models}                             & 29.1 & 61.4 & 37.1 & 22.7 & 37.6 & 28.7 & 60.6 & 37.2 & 21.8 & 37.1 \\
 & \methodname                                             & 37.1 & 65.7 & 45.7 & 24.5 & \textbf{43.3} & 36.2 & 64.8 & 45.2 & 24.1 & \textbf{42.6} \\
 & \methodname{} + VQ                                      & 25.9 & 49.6 & 26.6 & 27.6 & 32.4 & 25.8  & 49.8  & 27.1  & 28.0  & 32.7  \\
\bottomrule
\end{tabular}
\vspace{1em}
\caption{
  \textbf{Zero-shot downstream benchmark accuracy under pipeline-parallel communication compression.}
  Accuracy on HellaSwag, PIQA, ARC-Easy (ARC-E), and ARC-Challenge (ARC-C) for 150M, 500M, and 1B 
  parameter models trained with $P{=}4$ and $P{=}8$ pipeline stages.
  \methodname{} consistently approaches the uncompressed baseline across all benchmarks and scales—
  averaging within ${\sim}1$ point at 500M and 1B—while SSN baselines suffer larger accuracy drops 
  (up to ${\sim}9$ average points at 1B scale).
  The vector-quantized variant (\methodname{} + VQ) achieves higher compression ratios but at a 
  notable accuracy cost.
  Averages (Avg) are reported across all four tasks. The best compressed values marked with bold text.
}
\label{tab:downstream-accuracy}
\end{table}

\section{Conclusion}

We introduced \textbf{MAPL}, a communication-efficient pipeline-parallel training method that learns stage-specific orthogonal compression subspaces directly on the Stiefel manifold using manifold-constrained optimization. By combining adaptive low-rank projectors with lightweight factorized anchor embeddings and optional residual vector quantization, MAPL enables aggressive inter-stage activation compression while preserving training quality. Across LLaMA-style models ranging from 150M to 1B parameters and pipeline configurations with 4 and 8 stages, MAPL achieves $4\times$--$8\times$ communication compression within approximately $1$--$2\%$ of the uncompressed validation loss. It extends to $16\times$ compression with only modest additional degradation when combined with vector quantization. The method consistently outperforms recent work subspace networks~\citep{protocol_models} while maintaining strong downstream zero-shot accuracy. Our results suggest that inter-stage transformer activations naturally admit adaptive low-dimensional structure that can be learned jointly with the task objective, rather than imposed through a fixed global subspace. More broadly, this work highlights manifold-constrained representation learning as a practical mechanism for reducing communication costs in distributed foundation-model training.

\textbf{Limitations.}
Our experiments are limited to models up to 1B parameters, and further evaluation at larger scales and under real heterogeneous network conditions remains necessary. In addition, although MAPL substantially reduces the degradation associated with activation compression, performance still declines under extreme compression ratios, particularly when vector quantization is applied.

\bibliographystyle{plain}
\bibliography{ref}

%%%%%%%%%%%%%%%%%%%%%%%%%%%%%%%%%%%%%%%%%%%%%%%%%%%%%%%%%%%%
\clearpage
\appendix

\section{Background: Subspace networks}
\label{apx:subspace_background}
SSN~\citep{protocol_models} exploits a structural property of trained Transformers: the output projection weights $W_2^\ell \in \mathbb{R}^{d_{\mathrm{ff}} \times d}$ and attention projection weights $W_1^\ell \in \mathbb{R}^{d\times d}$ undergo \emph{rank collapse} during training, with their effective row space converging to a low-dimensional subspace $\mathcal{S} \subset \mathbb{R}^d$ of dimension $k \ll d$. This collapse arises naturally from AdamW's adaptive scaling, which attenuates gradient components aligned with negligible singular directions.

SSN operationalizes this observation via five steps. \textbf{(1)}~A fixed orthonormal basis $U_k \in \mathbb{R}^{d \times k}$ is constructed by QR decomposition of a random matrix. \textbf{(2)}~The rows of each $W_2^\ell$ and $W_1^\ell$ are constrained to $\mathrm{span}(U_k)$ by replacing the per-parameter adaptive scaling in AdamW with a row-wise constant variant. \textbf{(3)}~Deterministic components---positional embeddings $\mathrm{PE}$ and fixed token embeddings $T_{\mathrm{fixed}}$---are subtracted from $X^{b_p}$ before compression. \textbf{(4)}~The stage transmits the compressed activation
\begin{equation}
    Z^{b_p} = \bigl(X^{b_p} - \mathrm{PE} - T_{\mathrm{fixed}}\bigr)\, U_k \;\in\; \mathbb{R}^{B \times T \times k},
\end{equation}
achieving a compression ratio of $k/d$. \textbf{(5)}~The receiving stage reconstructs $\hat{X}^{b_p} = Z^{b_p} U_k^\top + \mathrm{PE} + T_{\mathrm{fixed}}$.

Because weight confinement ensures $X^{b_p} - \mathrm{PE} - T_{\mathrm{fixed}} \in \mathrm{span}(U_k)$ exactly, SSN's compression is \emph{architecturally lossless} on its modified model, though it incurs loss relative to an unconstrained baseline. The basis $U_k$ is refined via Grassmann manifold updates every ${\sim}500$ steps.

\textbf{Limitations.} SSN's losslessness is contingent on strict weight confinement. The subspace is fixed at initialization and only coarsely adapted; the model is forced to conform to a predetermined geometry rather than discovering a task-optimal one. 
In \S\ref{sec:method_obs}, we showed that pipeline boundary activations exhibit low-rank structure \emph{intrinsically}---even when trained with a modern optimizer~\citep{jordan2024muon} without any architectural constraint---motivating a jointly learned subspace that adapts to the activation geometry rather than imposing it.

\begin{table}[ht]
\small
\centering
\caption{Comparison of architectural features between Subspace Networks (SSN) and MAPL. Our method enables high-ratio compression without constraining model weights to low-rank subspaces, while maintaining compatibility with diverse optimizers through Stiefel manifold learning.}
\label{tab:method_comparison}
\begin{tabular}{lccc}
\toprule
\textbf{Feature} & \textbf{Subspace Networks (SSN)} & \textbf{MAPL (Ours)} & \textbf{MAPL + VQ} \\
\midrule
Unconstrained Full-Rank Weights & $\times$ & \checkmark & \checkmark \\
Learnable Projection Basis & Slow grassman updates & \checkmark & \checkmark \\
Decoupled low rank Embeddings & \checkmark & $\times$ & $\times$ \\
Learnable Per-Stage Anchor & $\times$ & \checkmark & \checkmark \\
Discrete Vector Quantization & $\times$ & $\times$ & \checkmark \\
\bottomrule
\end{tabular}
\end{table}

\section{Evaluating SSN without Subspace Projection}
\label{apdx:uk_ablation}

We ran probe evaluations on 16 sequences of 2048 tokens drawn from the DCLM validation set~\citep{dclm}. In this ablation, we replaced the subspace projection matrix $U_k$ with an identity matrix at inference time and replaced the decoupled low rank embedding table with its projections $TE \mathbin{@} U_k^\top$.

As a reference, the uniform-distribution baseline (i.e., random guessing over the vocabulary) yields a loss of $\approx 10.38$. Any value above this threshold indicates performance worse than chance.

The learned subspace projection matrices $U_k$ do more than compress information—they form an integral part of the model’s learned representations. When we replace them with the identity matrix at inference time (while keeping all weights trained with active projection), the model suffers severe degradation across every tested configuration. Losses increase by between $+3.44$ and $+8.25$, and several $P=8$ runs exceed the uniform baseline entirely. See Table \ref{tab:uk_ablation}.

This effect holds consistently across model scales (150M–1B), optimizers (Muon~\citep{jordan2024muon} and AdamW~\citep{adam}), and degrees of parallelism. These results show that the network internalizes representations tightly coupled to the projected subspaces. Removing the projection does not simply degrade the model but breaks it.

\begin{table}[ht]
\centering
\caption{%
  Loss with trained $U_k$ versus identity ablation ($U_k = I$).
  Values above $\approx 10.38$ (uniform random baseline) indicate worse-than-chance
  predictions.
}
\label{tab:uk_ablation}
\setlength{\tabcolsep}{6pt}
\begin{tabular}{cclccccc}
\toprule
\textbf{Size} & \textbf{pp} & \textbf{Optimizer} & $k$ & \textbf{Step}
  & \textbf{Loss (trained)} & \textbf{Loss ($U_k{=}I$)} & $\boldsymbol{\Delta}$ \\
\midrule
150M & 4 & Muon & 256 & 2{,}861 & 3.357 & 7.528 & $+4.17$ \\
150M & 4 & AdamW & 256 & 2{,}861 & 3.474 & 6.911 & $+3.44$ \\
150M & 8 & Muon & 256 & 2{,}861 & 3.371 & 11.588 & $+8.22$ \\
150M & 8 & AdamW & 256 & 2{,}861 & 3.489 & 11.138 & $+7.65$ \\
\midrule
500M & 4 & Muon & 256 & 9{,}535 & 3.061 & 9.043 & $+5.98$ \\
500M & 4 & AdamW & 256 & 9{,}535 & 3.213 & 8.432 & $+5.22$ \\
500M & 8 & Muon & 256 & 9{,}535 & 3.088 & 11.342 & $+8.25$ \\
500M & 8 & AdamW & 256 & 9{,}535 & 3.262 & 10.040 & $+6.78$ \\
\midrule
1B & 4 & Muon & 256 & 19{,}072 & 3.051 & 10.023 & $+6.97$ \\
1B & 4 & AdamW & 256 & 19{,}072 & 3.374 & 8.602 & $+5.23$ \\
1B & 8 & Muon & 256 & 19{,}072 & 3.081 & 11.108 & $+8.03$ \\
1B & 8 & AdamW & 256 & 19{,}072 & 3.361 & 11.510 & $+8.15$ \\
\bottomrule
\end{tabular}
\end{table}

\section{Keeping Projectors in the Stiefel Manifold}
\label{apx:manifold_escape}

In this section, we demonstrate the necessity of constraining subspace projectors to the Stiefel manifold during training. If the subspace projectors are trained naively, standard optimizer updates cause them to deviate from the manifold, thereby violating the orthonormality constraint. Consequently, the projection loses its isometric properties, allowing the model to learn representations outside the intended subspace.

To empirically validate this, we train a 150M parameter model under three distinct projection configurations. In the baseline setup, we initialize an orthogonal projector and freeze it for the duration of training. In the second configuration, the projector is updated using standard Euclidean gradients without any manifold constraints. As anticipated, these unconstrained updates cause the weight matrix to escape the manifold and lose its orthogonality, which actively degrades performance—yielding worse results than simply utilizing a fixed projection. Finally, enforcing the manifold constraint via SPEL~\citep{yang2026manifold} updates resolves this issue and achieves the lowest validation loss. These results are summarized in Table \ref{tab:projection_update_strategy}.

\begin{table}[h]
\centering
\caption{Impact of Stiefel manifold constraints on validation loss for a 150M parameter model. Unconstrained updates (Muon) degrade performance compared to a fixed projection, while SPEL updates successfully optimize the projector while maintaining orthogonality.}
\label{tab:projection_update_strategy}
\begin{tabular}{lcc}
\hline
\textbf{Optimizer} & \textbf{Val loss} & \textbf{Delta} \\
\hline
Fixed orthogonal (No updates) & 3.1673 & 1.19\% \\
Learnable orthogonal (Muon) & 3.2101 & 2.56\% \\
Learnable orthogonal + SPEL updates (SPEL) & \textbf{3.1564} & \textbf{0.84\%} \\
\hline
\end{tabular}
\end{table}

\section{Rank Collapse in Weight Matrices}
\label{apx:rank_collapse}

\begin{figure}[htbp]
    \centering
    \includegraphics[width=0.99\linewidth]{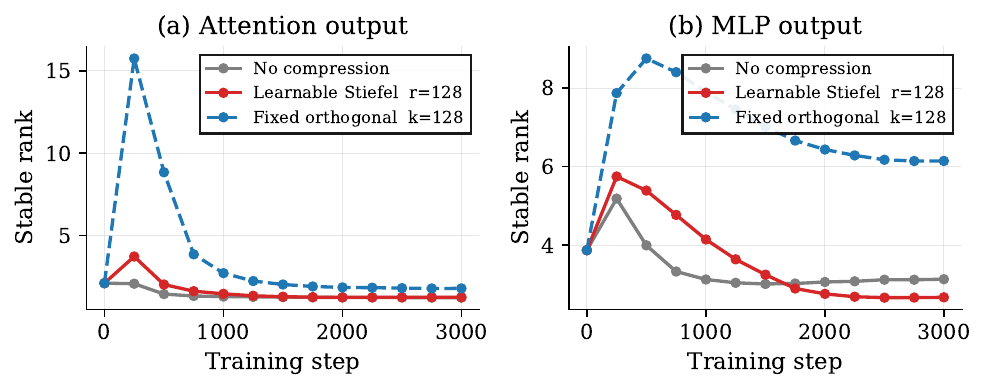}
    \caption{
        Dynamics of the stable rank, defined as $\mathrm{srank}(A)=\frac{\|A\|_F^2}{\|A\|_2^2}$, for layers adjacent to the compression boundary over the course of training. 
        (a) Under a learnable Stiefel compression ($r=128$), the attention output projection rapidly collapses to a severely low-rank structure, converging toward near rank-1 behavior. In contrast, a fixed orthogonal compression initially triggers a large transient increase in stable rank before partially collapsing. 
        (b) Similarly, the MLP output projection adapts toward a lower rank under the learned projector, whereas the fixed orthogonal basis maintains a substantially higher stable rank throughout training. 
        Together, these results demonstrate that jointly learning the projection matrix compels the network to reorganize its activations into a compact, low-dimensional subspace aligned with the communication rank. This explains the substantially higher residual energy retention achieved by the learned projector.
    }
    \label{fig:rank_collapse}
\end{figure}

Figure~\ref{fig:rank_collapse} illustrates the induced rank collapse observed when co-training the projection matrix on the Stiefel manifold. Learning the projection encourages the model to develop a low-rank structure that aligns closely with the learned basis. Consequently, the model exhibits a significantly more pronounced rank collapse compared to a fixed orthogonal projection. This structural adaptation accounts for the higher percentage of residual energy captured within the subspace. 

\section{Implementation Details}
\label{apx:implementation_details}

In this section, we provide a comprehensive overview of the models, datasets, and hyperparameters used across all experiments. Our goal is to ensure full reproducibility of the pipeline-parallel compression results presented in the main text. We ran our experiments of nodes consisting of 4$\times$ Nvidia H100 GPUs with precision bf16, we used Pytorch~\citep{paszke2019pytorch} framework with huggingface transformers~\citep{wolf2019huggingface} implementation of Llama 2~\citep{touvron2023llama}, with RoPE\citep{su2024roformer} positional embeddings. We used Python 3.10.  

\subsection{Model Architectures}
All experiments evaluate decoder-only transformers based on the LLaMA 2 architecture~\citep{touvron2023llama}. We define three scales of models: 150M, 500M, and 1B parameters. To ensure stable comparisons across scales, all models use a context sequence length of 2048 tokens and process inputs using a vocabulary size of 32,000. The structural hyper-parameters for each model scale are as follows:

\begin{itemize}
    \item \textbf{150M:} $d_{\mathrm{model}} = 1024$, $n_{\mathrm{layers}} = 9$, $n_{\mathrm{heads}} = 8$, $d_{\mathrm{ffn}} = 2688$
    \item \textbf{500M:} $d_{\mathrm{model}} = 1536$, $n_{\mathrm{layers}} = 18$, $n_{\mathrm{heads}} = 12$, $d_{\mathrm{ffn}} = 4096$
    \item \textbf{1B:} $d_{\mathrm{model}} = 2048$, $n_{\mathrm{layers}} = 20$, $n_{\mathrm{heads}} = 16$, $d_{\mathrm{ffn}} = 5440$
\end{itemize}

\subsection{Dataset and token budget}
Models were pre-trained on the DCLM dataset~\citep{dclm}. For all runs, we adhered to the Chinchilla compute-optimal scaling regime~\citep{hoffmann2022training}, defining our total training steps such that the models process 20 tokens per parameter. For example, the 150M model was trained on roughly 3 billion tokens, the 500M model on 10 billion tokens, and the 1B model on 20 billion tokens. A uniform 5M token split from DCLM was held out entirely for validation.

\subsection{Optimizer Configuration}
\label{apx:impl_optimizer}
We employed a hybrid optimizer approach. The 2D hidden weight matrices (attention projections and MLP matrices) were updated using the Muon optimizer~\citep{jordan2024muon}, using PolarExpress~\citep{amsel2025polar} algorithm for orthogonalizing the updates. The 1D parameters, including biases, layer normalizations, token embeddings, and the final language modeling head, were updated using AdamW~\citep{adamw}.

\textbf{Hyperparameter Sharing Strategy:} To ensure a fair comparison between the baseline (uncompressed) models and our proposed methods and baselines (MAPL, SSN, etc.), we used the exact same optimizer learning rates and weight decay as the uncompressed baselines across all compressed experiments, and we did not tune them specifically for the compression techniques. 
For the main model parameters, we set the Muon learning rate to $\eta_\mu = 0.02$ and the AdamW learning rate to $\eta_{\mathrm{adam}} = 0.01$ (a $0.5\times$ multiplier on the Muon rate). For the 1B model experiments , we used $\eta_\mu = 0.01$ and $\eta_{\mathrm{adam}} = 0.005$. We use a global batch size of 512 for all runs. 

\subsection{MAPL Compression Hyperparameters}
Our method relies on learning an orthogonal projector $A_p \in \mathrm{St}(d, r)$ per pipeline stage boundary. We optimized these Stiefel manifold projectors using the manifold constrained steepest descent algorithm~\citep{yang2026manifold} via a Polar Express retraction~\citep{amsel2025polar}. We found empirically that the projector optimization benefits from a slightly reduced learning rate compared to the main model weights; therefore, we set the projector learning rate to be $0.1\times$ the main Muon learning rate.

For the Vector Quantized (MAPL + VQ) experiments, we used a codebook size of $K=256$ and applied $R=2$ residual quantization rounds with a group dimension of $G=2$, utilizing an asynchronous streaming dictionary of steaming interval $5$, to amortize codebook synchronization costs across micro-batches.

\section{Amortizing VQ: Streaming Codebook Synchronization}
\label{apx:vq_streaming}

For our vector-quantized variant (MAPL + VQ), both the sender and receiver stages must maintain identical codebooks. Transmitting the full codebook $\mathcal{C}_p \in \mathbb{R}^{r \times K}$ across the network at every micro-batch would severely erode the bandwidth savings achieved by quantization. 

Motivated by the observation that VQ codebooks evolve slowly over training~\citep{zhang2024codebook}, we introduced an asynchronous streaming dictionary update protocol (\S\ref{sec:method_vq}). Instead of full synchronization, the sender transmits only a small, randomly sampled subset of the codebook (e.g., $1/K$ of the codes) alongside the quantized activations in each micro-batch.

To validate that this partial synchronization does not degrade representational fidelity, we conducted an ablation comparing full codebook synchronization against our streaming protocol on a 150M model ($P=4$, $8\times$ compression). The model trained with \textbf{full synchronization} achieved a validation loss of $3.1647$, while the model trained with the \textbf{streaming dictionary protocol} achieved an essentially identical validation loss of $3.1651$. This result confirms that continuous, fractional codebook updates are sufficient to maintain alignment between pipeline stages without incurring the communication penalty of full synchronization.

\section{Ablation of Factorized Anchors}
\label{sec:anchor_ablation}

In Section~\ref{sec:method_construction}, we hypothesized that the token-driven offset in the residual stream is inherently high-rank and consumes projector capacity if not explicitly modeled. To validate this, we ablate the anchor formulation on a 150M model across an 8-stage pipeline ($P=8$, $4\times$ compression), a depth where compression errors severely compound.

As shown in Table~\ref{tab:anchor_ablation}, removing the anchor entirely degrades the validation loss to $3.209$ (a $+2.39\%$ degradation relative to the uncompressed baseline). Adding a static, frozen embedding projection (analogous to the static offset in SSN) fails to improve performance ($3.212$). Conversely, utilizing a fully trainable per-stage anchor recovers performance to $3.149$, almost matching the uncompressed baseline ($3.134$), but incurs an unacceptable parameter overhead (an additional full embedding table per stage).

Our proposed \emph{factorized anchor} strikes the optimal balance: by learning a low-rank trainable component ($E_p^{\mathrm{small}}$) against a frozen random projection ($P_p$), it recovers the majority of the performance ($3.165$, only $+1.11\%$ degradation) while adding negligible parameter overhead to the pipeline stages.

\begin{table}[ht]
\centering
\caption{Ablation of anchor embedding strategies (150M model, $P=8$, 4$\times$ compression). The factorized anchor recovers performance comparable to a fully trainable anchor but with negligible parameter overhead.}
\label{tab:anchor_ablation}
\begin{tabular}{llcc}
\toprule
\textbf{Anchor Strategy} & \textbf{Validation Loss} & \textbf{$\Delta$ \% vs Baseline} \\
\midrule
\textit{Uncompressed Baseline ($P=8$)} & 3.134 & --- \\
\midrule
No Anchor & 3.209 & +2.39\% \\
Decoupled embedding (SSN~\citep{protocol_models} style) & 3.212 & +2.48\% \\
Full Trainable Anchor & \textbf{3.149} & \textbf{+0.47\%} \\
\textbf{Factorized Anchor (Ours)} & 3.165 & +1.11\% \\
\bottomrule
\end{tabular}
\end{table}

% \subsection{Depth Scaling and Error Accumulation}
% \label{sec:depth_scaling}

% Unlike data-parallel compression, where errors are averaged across workers, pipeline-parallel compression errors compound sequentially across stage boundaries. We evaluate the robustness of MAPL as the pipeline depth $P$ increases from 4 to 16 stages on the 500M parameter model at an $8\times$ compression ratio.

% When employing learned orthogonal projectors \emph{without} anchor embeddings, the compounding error becomes catastrophic at depth: degradation jumps from $+1.56\%$ at $P=4$ to substantial divergence at $P=16$. However, when equipped with factorized anchor embeddings, MAPL effectively arrests this accumulation. At $P=8$, the method matches the uncompressed baseline (2.837 vs 2.835), demonstrating that the per-stage anchors successfully absorb the compounding high-rank token offsets, allowing the Stiefel projectors to losslessly route the task-specific signal even through highly fragmented pipelines.

\section{Summary of Hyperparameters}
\begin{table}[h]
\centering
\small
\caption{\textbf{Hyperparameters shared across all methods and scales.}}
\label{tab:shared_hparams}
\begin{tabular}{@{}ll@{}}
\toprule
\textbf{Hyperparameter}      & \textbf{Value} \\
\midrule
Global batch size            & 512 \\
Micro-batch size             & 4 \\
Sequence length              & 2048 \\
Pipeline-parallel degree $P$ & $\{4, 8\}$ \\
Precision                    & bf16 \\
Attention implementation     & SDPA \\
Muon momentum                & 0.95 \\
Weight decay            & 0.01 \\
Gradient clipping            & off (1.0 for VQ runs at 1B) \\
Validation split             & 5M tokens, held-out from DCLM~\citep{dclm} \\
\bottomrule
\end{tabular}
\end{table}

\begin{table}[h]
\centering
\small
\setlength{\tabcolsep}{4pt}
\caption{\textbf{Method-specific hyperparameters for the runs in Table~\ref{tab:my-table}.} ``LR scale'' is $\eta_{\mathrm{adam}}/\eta_\mu$, the AdamW-to-Muon learning-rate multiplier on the $1$D parameters (embeddings, biases, LayerNorms, LM head). The SSN baselines update the shared subspace $U_k$ via Grassmann manifold steps every $500$ optimizer steps with a fixed Grassmann learning rate of $0.01$. ``Compression'' values are reported relative to a $2$-byte (bf16) per-channel payload at the corresponding hidden dimension; SSN, MAPL, and MAPL+VQ are all rank-matched at $r{=}256$. Identical values across the three model scales are denoted ``--''.}
\label{tab:method_hparams}
\begin{tabular}{@{}llccc@{}}
\toprule
\textbf{Method} & \textbf{Hyperparameter} & \textbf{150M} & \textbf{500M} & \textbf{1B} \\
\midrule
\multirow{2}{*}{\textit{Uncompressed}}
  & Optimizer (2D / 1D)                                & Muon / AdamW & -- & -- \\
  & LR scale ($\eta_{\mathrm{adam}}/\eta_\mu$)         & 0.5          & -- & -- \\
\midrule
\multirow{4}{*}{SSN (AdamW)}
  & Subspace rank $k$                                  & 256          & -- & -- \\
  & Grassmann update period / LR                       & 500 steps / $1{\times}10^{-2}$ & -- & -- \\
  & Optimizer                                          & AdamW        & -- & -- \\
  & AdamW learning rate                                & $3{\times}10^{-3}$ & $3{\times}10^{-3}$ & $2{\times}10^{-3}$ \\
\midrule
\multirow{3}{*}{SSN + Muon}
  & Subspace rank $k$                                  & 256          & -- & -- \\
  & Grassmann update period / LR                       & 500 steps / $1{\times}10^{-2}$ & -- & -- \\
  & LR scale ($\eta_{\mathrm{adam}}/\eta_\mu$)         & 0.1          & -- & -- \\
\midrule
\multirow{6}{*}{\methodname}
  & Projection rank $r$                                & 256          & -- & -- \\
  & Compression ratio (vs.\ bf16)                      & $4\times$    & $6\times$ & $8\times$ \\
  & Anchor rank ($E_p^{\mathrm{small}}\!\in\!\mathbb{R}^{V\times r}$) & 256 & -- & -- \\
  & Anchor projector $P_p$                             & frozen, random orthonormal & -- & -- \\
  & SPEL LR multiplier $\alpha/\eta_\mu$               & 0.1          & -- & -- \\
  & SPEL retraction (LMO / projection steps)        & 5 / 7, Polar Express~\citep{amsel2025polar} & -- & -- \\
\midrule
\multirow{5}{*}{\makecell[l]{\methodname \\ + VQ}}
  & All MAPL hyperparameters                        & (as above) & -- & -- \\
  & Codebook size $K$ / residual rounds $R$            & 256 / 2     & -- & -- \\
  & VQ group size $G$                                  & 2            & -- & -- \\
  & Streaming dictionary refresh                       & $1/5$ codes per micro-batch & -- & -- \\
  & Compression ratio (vs.\ bf16)                      & $8\times$   & $12\times$ & $16\times$ \\
\bottomrule
\end{tabular}
\end{table}

\clearpage

\end{document}